\documentclass[a4paper, 10pt, conference]{IEEEtran}
\IEEEoverridecommandlockouts
\usepackage{cite}
\usepackage{amsmath,amssymb,amsfonts}
\usepackage{algorithmic}
\usepackage{graphicx}
\usepackage{textcomp}
\usepackage{xcolor}
\def\BibTeX{{\rm B\kern-.05em{\sc i\kern-.025em b}\kern-.08em
    T\kern-.1667em\lower.7ex\hbox{E}\kern-.125emX}}
    
\usepackage{epsfig}
 
\usepackage{array}
\usepackage{multirow}
\usepackage{adjustbox}
\usepackage{booktabs,siunitx}
\usepackage{graphicx}
\usepackage{subfigure}
\usepackage{amsmath}
\usepackage{xcolor}
\usepackage{color, soul}
\usepackage{stfloats}
\usepackage{amsfonts,amssymb}
\usepackage{algorithm}
\usepackage{algorithmic}
\usepackage{url}
\usepackage{enumitem}
\usepackage{interval}

\usepackage{threeparttable}
\usepackage{booktabs}
\usepackage{verbatimbox}

\begin{document}
\title{Fusion of Physiological and Behavioural Signals on SPD Manifolds with Application to Stress and Pain Detection\\
\thanks{The proposed work was supported by the French State, managed by the National Agency for Research (ANR) under the Investments for the future program with reference ANR-16-IDEX-0004 ULNE.}
}
\makeatletter
\newcommand{\linebreakand}{%
  \end{@IEEEauthorhalign}
  \hfill\mbox{}\par
  \mbox{}\hfill\begin{@IEEEauthorhalign}
}
\makeatother
\author{\IEEEauthorblockN{Yujin WU}
\IEEEauthorblockA{
\textit{CNRS, UMR 9189 CRIStAL}\\
\textit{University of Lille, Centrale Lille}\\
Lille, France 59000\\
yujin.wu.etu@univ-lille.fr}
\and
\IEEEauthorblockN{Mohamed Daoudi}
\IEEEauthorblockA{
\textit{IMT Nord Europe, Institut Mines-Télécom, Centre for Digital Systems}\\
\textit{University of Lille, Centrale Lille, CNRS, UMR 9189 CRIStAL}\\
Lille, France 59000\\
mohamed.daoudi@imt-nord-europe.fr}
\linebreakand
\IEEEauthorblockN{Ali Amad}
\IEEEauthorblockA{
\textit{LilNCog, U1172 INSERM}\\
\textit{University of Lille} \\
Lille, France 59000\\
ali.amad@univ-lille.fr}
\and
\IEEEauthorblockN{Laurent Sparrow,}
\IEEEauthorblockA{
\textit{CNRS, UMR 9193 SCALab}\\
\textit{University of Lille} \\
Lille, France 59000\\
laurent.sparrow@univ-lille.fr}
\and
\IEEEauthorblockN{Fabien~D'Hondt}
\IEEEauthorblockA{
\textit{LilNCog, U1172 INSERM}\\
\textit{University of Lille} \\
Lille, France 59000\\
fabien.d-hondt@univ-lille.fr}
}
\maketitle
\begin{abstract}
Existing multimodal stress/pain recognition approaches generally extract features from different modalities independently and thus ignore cross-modality correlations. This paper proposes a novel geometric framework for multimodal stress/pain detection utilizing Symmetric Positive Definite (SPD) matrices as a representation that incorporates the correlation relationship of physiological and behavioural signals from covariance and cross-covariance. Considering the non-linearity of the Riemannian manifold of SPD matrices, well-known machine learning techniques are not suited to classify these matrices. Therefore, a tangent space mapping method is adopted to map the derived SPD matrix sequences to the vector sequences in the tangent space where the LSTM-based network can be applied for classification. The proposed framework has been evaluated on two public multimodal datasets, achieving both the state-of-the-art results for stress and pain detection tasks.
\end{abstract}

\begin{IEEEkeywords}
stress detection, pain detection, multimodal fusion, covariance matrix, symmetric positive definite manifold.
\end{IEEEkeywords}

\section{Introduction}
Among a wide range of applications for monitoring and controlling human physical and mental health, stress/pain detection has drawn massive attention.  
Stress is a set of complex physiological, cognitive, and behavioral responses that are triggered by a challenging condition. 
When exposed to several stressors over a long period of time, a person's mental and physical state can be negatively influenced, which can further result in chronic health problems\cite{Gedam_review_mentalStress}. 
In order to detect stress at an early stage to reduce its impact on health conditions, stress detection has been widely used in many fields such as automated driver assistance\cite{stress_driver}, academic environment\cite{stress_academic}, social communication\cite{stress_social}, etc. 
On the other hand, pain is another unpleasant sensation that we may encounter in our daily lives. 
Its accurate assessment plays a key role in diagnosing the condition, monitoring post-operative progress and optimising treatment options\cite{Werner_pain}.
Thus, automated pain detection is another emerging area of healthcare applications designed to precisely assess pain to mitigate subject errors associated with patient self-reporting. 
A variety of sensors can be applied to collect stress/pain indicators from different perspectives, whereby the corresponding approaches for stress/pain detection can be mainly divided into three categories: 1) physiological based methods via bio-signals (e.g., electroencephalogram (EEG), electrodermal activity (EDA), etc.); 2) behavioural-based methods via physical signals (e.g., facial expressions, speech, body movements, etc.); 3) multimodal-based methods via a combination of physiological and behavioural signals. As complementary information between multimodalities contributes more to the robustness and reliability of the system, therefore, stress/pain detection combining physiological and behavioural indicators has become more attractive. 
However, how to effectively fuse multimodal data remains an important challenge for such systems. Most existing multimodal-based approaches learn features independently in each modality and eventually integrate them at the feature level or decision level.  Thus underlying correlations between multiple modalities are ignored. 
In this study, we address the above problem by introducing a geometric framework that employs symmetric positive definite (SPD) matrices extracted from physiological and behavioural signals as a joint multimodal feature representation on SPD manifold for stress/pain detection.
Continuous multimodal data recording can first be converted into SPD matrix sequences. The tangent space mapping method is then applied to locally flatten the manifold and approximate the SPD matrix sequences by tangent space vector sequences, where an LSTM-based deep neural network can be implemented to learn the context correlations of input sequences for classification. The overview of the proposed method is illustrated in Fig. \ref{Fig1:Overview}.
To the best of our knowledge, this is the first exploration of applying the SPD matrix to fuse behavioural signals and physiological signals for stress/pain detection.

\section{Related Work}
\label{sec:Related Work}
The multimodal framework is promising to improve the performance of stress/pain detection. However, few studies can be found using combined data from different fields (i.e. physiological and behavioural).
\paragraph*{Multimodal data fusion}
The multimodal fusion techniques can be divided into three categories: early, intermediate and late fusion. In an early fusion, features acquired by different sensors were fused into a unique representation before performing a learning task. Compared with the other two fusions, early fusion is widely explored in literature. 
Aigrain et al.\cite{Aigrain_stress} captured body video, high-resolution facial images and physiological signals from 25 subjects during a mental arithmetic test for stress detection, where 101 features from behavioural and physiological signals were extracted to train an SVM classifier. Rastgoo et al. \cite{RASTGOO_stress} proposed a multimodal fusion framework for driver stress detection where the ECG signal, vehicle data, and contextual data were each entered into separate CNNs to extract modality-specific features. The LSTM-based network was trained on the concatenated multimodal feature embeddings to detect stress levels.
Werner et al. \cite{multimodal_pain_Werner} collected facial
distances and gradient-based features from video frames which were combined with the statistical features calculated from biological signals. The resulting multimodal vectors were employed to train a random forest model for pain assessment.
The multimodal fusion approaches described above show promising results for stress/pain detection. However, when simply splicing features from each modality for classification, only correlations within individual modalities were considered and the potential of applying inter-modal interactions to further boost performance is neglected. 

\paragraph*{Symmetric Positive Definite (SPD) matrices} 
Recently, covariance-based representations have gained great popularity in computer vision and machine learning. This success can be explained by three major advantages. 
Firstly, several characteristics can be fused into a single tensor and deliver higher-order statistical information. Secondly, the covariance matrices are symmetric positive definite (SPD) matrices with well-established mathematical theoretical properties~\cite{bharia2009,zhang_cross-covariance_2020}.
In addition, the SPD matrices have shown impressive accomplishments in many real-world applications such as pedestrian detection ~\cite{TuzelPami2008}, facial expression recognition ~\cite{OtberdoutIEEENNLS2020}, brain-computer interfaces ~\cite{barachant_classification_2013}, etc. However, all of these applications concentrate solely on behavioural perspectives or physiological perspectives.
Liu et al.\cite{liu_spdmulti} proposed a multimodal emotion recognition approach using video and audio modalities. Covariance matrix, linear subspace, and Gaussian distribution were built from facial video frames and regarded as points on Riemannian manifolds. Subsequently, the similarity matrix calculated using different Riemann kernels is fed into multiple classifiers (i.e., SVM, partial least squares, and logistic regression). However, they only constructed the SPD matrix for the video modality, the audio features were extracted using an existing toolkit, and fed into the same type of classifier as the video modality.
In the end, the final fusion of the two modalities was established on the decision level. Thus, the exploration of inter-modal correlations using SPD matrix is still missing here.
In the work of \cite{zhang_cross-covariance_2020}, they presented a more general covariance-based SPD representation, containing additional cross-covariance information from different time steps for action recognition. Inspired by \cite{zhang_cross-covariance_2020}, we migrate this new type of tensor representation to the scenario of multimodal stress/pain detection and show its effectiveness in this paper. Different from the previous work, we fuse behavioural and physiological information into one single SPD matrix-based representation, which not only incorporates intra-modal correlations, but also allows for exchanges across two modalities.
To the best of our knowledge, this is the first use of the geometry of SPD manifold matrices to merge physiological and behavioural signals.
\begin{figure*}[!htb]
\centering
\includegraphics[width=6in]{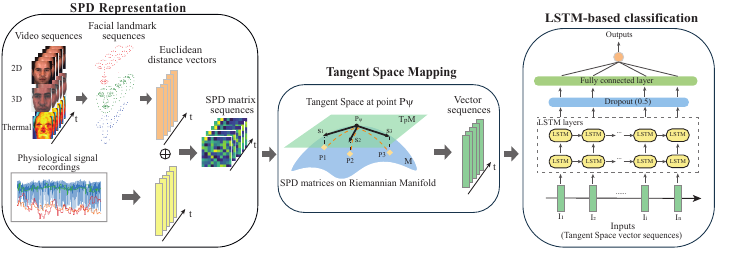}
\caption{Overview of the proposed framework. First, the SPD matrix sequences that incorporate the correlation information between multimodal data (i.e. physiological and behavioural signals) can be extracted from the segmented data records. Subsequently, the tangent space mapping projects the SPD matrix sequences to the vector sequences in the tangent space. Finally, these vectors can be used as input to the LSTM-based classification network for stress/pain recognition.}
\label{Fig1:Overview}
\end{figure*}
\section{Proposed Method}
\label{sec:Proposed Method}
\subsection{Symmetric positive definite (SPD) matrix for multimodal signal}
Let $\mathbf{x_i}=[v_{1},...,v_{D}]^T\in\mathbb{R}^{D}, D\geq2$ represent a multimodal signal vector comprising the behavioural and physiological signals at the i-th timestamp, where the number of signals is denoted by D. A short-segment can be extracted from the continuous signal recordings of a trial, resulting in a centered matrix $ \mathbf{X}$ = $[\mathbf{x_{1}},..., \mathbf{x_{i}}...,\mathbf{x_{N}}]\in\mathbb{R}^{D \times N}$, where N is the number of time instants for each segment. The outer product operation (denoted by the symbol $\otimes$) is then performed on all signal column vector pairs ($\mathbf{x_i},\mathbf{x_j}$) for i and j $= 1,...,N$ in $\mathbf{X}$ and consequently produces a partition matrix :
\begin{equation}
  \mathbf{\Omega}=\left[
  \begin{matrix}
  \setlength{\arraycolsep}{0.7pt}
   \mathbf{x_{1}} \otimes \mathbf{x_{1}}  & \cdots & \mathbf{x_{1}} \otimes \mathbf{x_{N}}
   \\\mathbf{x_{2}} \otimes \mathbf{x_{1}} & \cdots & \mathbf{x_{2}} \otimes \mathbf{x_{N}}
   \\\vdots & \ddots & \vdots \\
   \mathbf{x_{N}} \otimes \mathbf{x_{1}}  & ... & \mathbf{x_{N}} \otimes \mathbf{x_{N}}
  \end{matrix}
  \right] \in \mathbb{R}^{DN \times DN}
  \label{eq:1}
\end{equation}
where the element of $\mathbf{\Omega}$ at position (i, j) is given by:
\begin{equation}
  \mathbf{\Omega(i,j)} = \mathbf{x_{i}} \otimes \mathbf{x_{j}} = \mathbf{x_i}\mathbf{x_j}^{T} \in \mathbb{R}^{D \times D}
  \label{eq:2}
\end{equation}
The sample covariance matrix $\mathbf{S}$ which is a SPD matrix can be derived from the diagonal elements of $\mathbf{\Omega}$:
\begin{equation}
  \mathbf{S} = \frac{1}{N-1}\sum_{i=1}^N  \mathbf{\Omega(i,i)} = \frac{1}{N-1}\sum_{i=1}^N \mathbf{x_{i}}\mathbf{x_{i}}^{T} \in \mathbb{R}^{D \times D}
  \label{eq:3}
\end{equation}
The second SPD matrix defined as cross-covariance can be extracted from the off-diagonal elements of $\mathbf{\Omega}$, which contains the correlation information of the signal vectors at different timestamps and is denoted by $\mathbf{C}$:
\begin{equation}
\begin{aligned}
\mathbf{C}&= \frac{1}{N^2 - N}\sum_{i=1,j = 1,i\ne j}^{N,N}  \mathbf{\Omega(i,j)} \\
&=\frac{1}{N^2 - N}\sum_{i=1,j = 1,i\ne j}^{N,N} \mathbf{x_{i}}\mathbf{x_{j}}^{T}\in \mathbb{R}^{D \times D}
\end{aligned}
\label{eq:4}
\end{equation}

The covariance $\mathbf{S}$ and cross-covariance $\mathbf{C}$, are then combined in a symmetric manner to form a new block matrix which remains a SPD matrix and is denoted by $\mathbf{P}$:
\begin{equation}
  \mathbf{P}
  =\left[
    \begin{matrix}
    \mathbf{S} &\mathbf{C} &\cdots &\mathbf{C}\\
    \mathbf{C} &\mathbf{S} &\cdots &\mathbf{C}\\
    \vdots & \vdots & \ddots & \vdots\\
    \mathbf{C} &\mathbf{C} &\cdots &\mathbf{S}\\
    \end{matrix}
    \right] \in \mathbb{R}^{(m\times D) \times (m\times D)}
    \label{eq:5}
\end{equation}
where m is the dimension of the new SPD matrix $\mathbf{P}$, in other words, $\mathbf{P}$ is composed of m blocks of $\mathbf{S}$ and $m(m-1)$ blocks of $\mathbf{C}$. A larger m corresponds to a higher computational cost, while the information ratio of the covariance to the cross-covariance decreases \cite{zhang_cross-covariance_2020}.


\subsection{Riemannian Geometry of symmetric positive definite (SPD) Matrices }
\label{sec:RiemannainGeo}
\paragraph*{Tangent Space Mapping}
A number of metrics have been proposed for SPD matrices. In this study, we consider one of the most popular metrics, the log-Euclidean metric~\cite{arsigny_logeuclidean:2006}, for its simplicity to compute. 
Let $\mathbf{P_{i}} $ and  $\mathbf{P_{j}}$ be any two points on the SPD manifold then the log-Euclidean metric is defined as:
\begin{equation}
  \delta_{R}(\mathbf {P_{i}}, \mathbf {P_{j}}) ={\lVert \mathrm{log} (\mathbf{P_{i}}^{-1}\mathbf{P_{j}})\rVert}_{F}
  \label{eq:6}
\end{equation}
where ${\lVert \cdot \rVert}_{F}$ is the Frobenius norm operator, and $\mathrm{log}(.,.)$ is the matrix log operateur.
The geometric mean\cite{barachant_multiclass_2012, moakher2005a} of a set of SPD matrices $ \{\mathbf{P_{1}},\mathbf{P_{2}},\ldots, \mathbf{P_{I}}\}, I\geq1$, $\mathbf{P_{i}} \in \mathcal{P}(n)$ can be derived using the  Riemannian geodesic distance \eqref{eq:6}:
\begin{equation}
  \psi(\mathbf{P_{1}},\mathbf{P_{2}},\ldots, \mathbf{P_{I}}) = \mathop{\arg\min}_{\mathbf{P} \in P(n)} \sum_{i=1}^{I} {\delta_{R}}(\mathbf{P}, \mathbf{P_{i}})^{2}
  \label{eq:7}
\end{equation}
The tangent space at $\mathbf{P} \in \mathcal{M}$, denoted by $T_{\mathbf{p}} \mathcal{M}$, is a real vector
space containing all tangent vectors to $\mathcal{M}$ at $\mathbf{P}$. The exponential map $\operatorname{Exp}_{\mathbf{P}}(\cdot): T_{\mathbf{P}} \mathcal{M} \rightarrow \mathcal{M}$ and its inverse, the
logarithm map, $\operatorname{Log_{\mathbf{P}}}(\cdot): \mathcal{M} \rightarrow T_{\mathbf{P}} \mathcal{M}$ are defined over Riemannian manifolds for exchanging between the manifold and its tangent space at $\mathbf{P}$. A tangent vector can be mapped to a point on the manifold through the exponential operator. The logarithm operator can project a point on the manifold to the tangent space\cite{Faraki_2015}. In the work of  \cite{barachant_multiclass_2012}, tangent space mapping was introduced which employs the tangent space that lies at the geometric mean \eqref{eq:7} of the 
entire set of SPD matrices: $\mathbf{P}_{\psi} =\psi(\mathbf{P_i},i=1,\ldots,I)$. Each SPD matrix $\mathbf{P_i}$ can be projected into the tangent space as a vector  $s_{i}$ with a dimension of $\frac{n(n+1)}{2}$:
\begin{equation}
 s_{i} = vec({\mathbf{P}_{\psi}}^{-\frac{1}{2}}\mathrm{Log}_{\mathbf{P_{\psi}}}(\mathbf{P_{i}}) {\mathbf{P_{\psi}}}^{-\frac{1}{2}}) \in\mathbb{R}^{\frac{n(n+1)}{2}}
  \label{eq:8}
\end{equation}
where the vector operator of one SPD matrix $\mathbf{P}$ is defined as
\begin{equation}
\resizebox{.9\hsize}{!}{
 $vec(\mathbf{P}) = [p_{1,1}, \sqrt{2}p_{1,2}, \sqrt{2}p_{1,3}, \cdots, p_{2,2}, \sqrt{2}p_{2,3}, \cdots, p_{n,n}]$}
  \label{eq:9}
\end{equation}
$p_{i,j} \in \mathbf{P}$ is the element of $\mathbf{P}$. This operation is designed to achieve vectorization by weighting the upper triangular part of the matrix, where the diagonal elements and off-diagonal elements  are multiplied by the unit weights and weights of $\sqrt{2}$, respectively\cite{tuzel_pedestrian_2008}.

\subsection{Classification of SPD matrix sequences }
Let us consider a set
$\mathcal{L}=\bigcup\limits_{q=1}^{Q}  \mathcal{L}^q$
consisting of data from Q subjects. The qth subset $\mathcal{L}^q$ is represented by $\mathcal{L}^q = \left\{([\mathbf{P_i},...,\mathbf{P_{i+T}}] , {y_i}), \mathbf{P_i} \in \mathcal{P}(n), i\in\interval{1}{|\mathcal{L}^q|} \right\}$
,where $[\mathbf{P_i},...,\mathbf{P_{i+T}}]$ is a segmented sequence consisting of T subsequences, $\mathbf{P_i}$ can be considered as a representation of the correlation information for the corresponding subsequence living on the manifold, and $y_{i}$ is the stress/pain label associated with the entire SPD matrix sequence, such that $y_{i}=f\left([\mathbf{P_i},...,\mathbf{P_{i+T}}]\right)$. 
In current literature, few approaches have been suggested to tackle the non-linearity of the SPD manifold. A common method of dealing with the non-linearity is to estimate the manifold-value data by projecting them into the tangent space of a specific point on the manifold (e.g., the geometric mean of the data)\cite{JayasumanaPAMI2015}.
In this work, we will adopt this approach to solve our problem. 
For each subset, its corresponding geometric mean ${\mathbf{P}_{\psi}}^q$ can be obtained with the equation \eqref{eq:7}. Then each SPD matrix $\mathbf{P_i}$ in the qth subset $\mathcal{L}^q$ is mapped into the tangent space and the derived corresponding subset of vector sequences is denoted by $s^q = \left\{([{s_i},..., s_{i+T}], {y_i}), s_i \in \mathbb{R}^{\frac{n(n+1)}{2}}, i\in\interval{1}{|s^q|} \right\}$ using the equation \eqref{eq:8}. The above process will be repeated for each subject's data. 
In the end, $\mathcal{L^{\ast}}=\bigcup\limits_{q=1}^{Q}  s^q$
is considered as the input of the LSTM-based deep neural network in Fig. \ref{Fig1:Overview}. During training, the temporal contextual relationships within the tangent space vector sequence are explored by the 2-layer LSTM network, and the output features are then fed into the fully connected layer, followed by the sigmoid function to obtain the predicted probabilities. 
\begin{figure*}[!htb]
\centering
\includegraphics[scale=0.46]{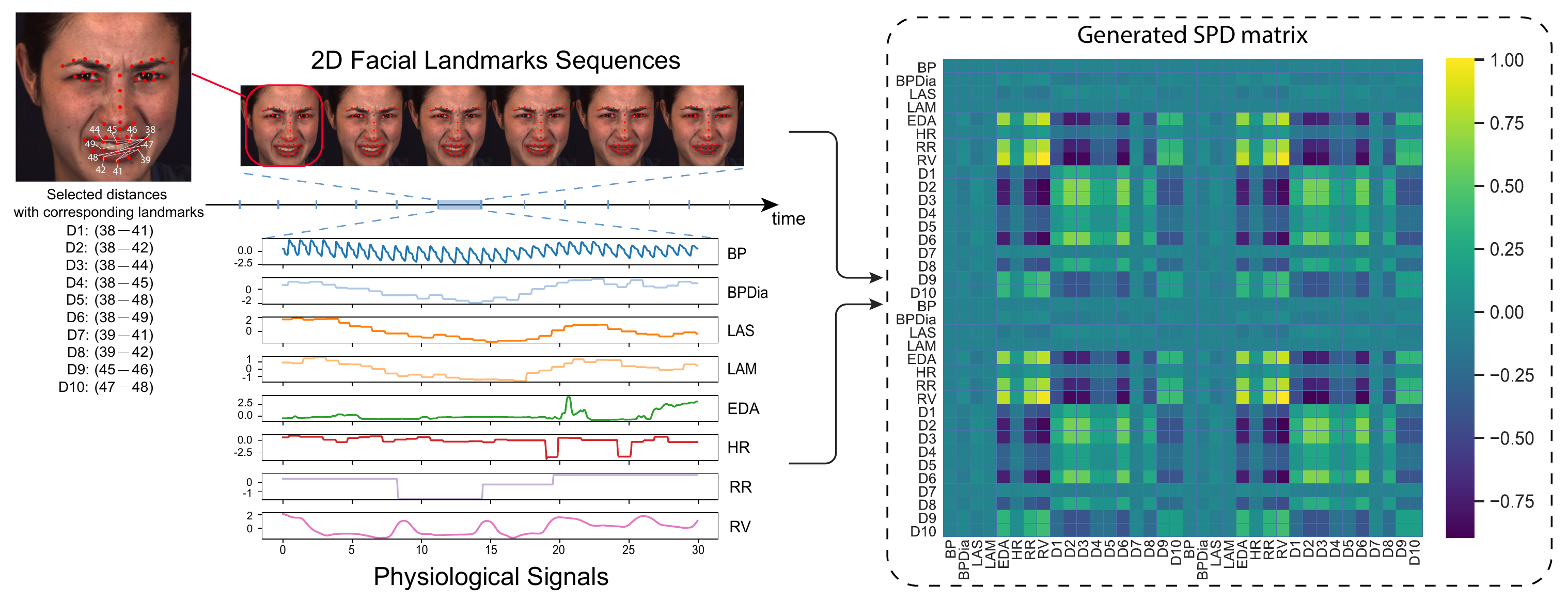}
\caption{The multimodal SPD representation generated by a pain sample in the BP4D+ dataset, where correlations within and across two modalities (i.e., vision and physiology) can be observed. (D1-D10: 10 distances automatically selected based on Anova F-value, BP: raw blood pressure, BPDia: diastolic blood pressure, LAS: systolic blood pressure, LAM: mean blood pressure, EDA: electrodermal activity, HR: heart rate, RR: respiration rate, and RV: respiration volts.)}
\label{Fig2:spd matrix}
\end{figure*}

\section{Experiments and Results}
\label{sec:Experiments and Results}
To evaluate the validity of the proposed method, we conducted multimodal stress detection experiments on WESAD dataset while multimodal pain detection experiments were carried out on BP4D+ dataset.
\subsection{Datasets}
The \textbf{WESAD dataset} \cite{schmidt_introducing_2018} is one of the most widely used public datasets for stress and affect recognition. In a restricted laboratory setting, multimodal data consisting of motion and physiological signals from 15 subjects were captured by two wearable devices, a wrist sensor and a chest sensor, respectively, and the experimental protocol was designed to stimulate three different emotional states (baseline, stress, amusement) in the participants. Based on previous work ~\cite{schmidt_introducing_2018,StressNAS,Lai_stress}, a binary stress detection problem (\textit{stress} vs. \textit{non-stress}) can be formulated on the WESAD dataset by fusing baseline class and amusement class to form the \textit{non-stress} class.

The \textbf{BP4D+ dataset}\cite{zhang_multimodal_2016} is a large-scale multimodal spontaneous emotion database. 140 subjects were required to complete 10 tasks to elicit 10 different emotions, during which 2D RGB images, 3D model sequences, thermal videos and 8 physiological signal sequences with 1.5 million frames were captured by different sensors. In addition, the metadata are also provided, including 2D/3D/thermal facial landmarks, hand-labelled FACS codes and auto-tracked head poses.
In this work, we focus on the identification of pain. As the dataset only provides the most facially-expressive segments for four emotions: happy, embarrassment, fear and pain, we therefore performed a pain detection task (\textit{pain} vs. \textit{non-pain}) by combining happy, embarrassment, fear as the \textit{non-pain} class as proposed in \cite{hinduja_multimodal_nodate}. An example of 2D texture images/3D model sequences/thermal images from the \textit{Pain} class and their corresponding facial landmarks is shown in Fig. \ref{Fig2: bp4d_example}.

\begin{figure}[htbp]
\centering
\includegraphics[scale = 1.6]{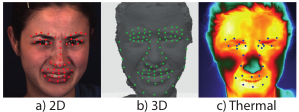}
\caption{An example of 2D texture images/3D model sequences/thermal images from the \textit{Pain} class and their corresponding facial landmarks provided in BP4D+ dataset.}
\label{Fig2: bp4d_example}
\end{figure}
\subsection{Data Preprocessing and SPD matrix construction}
For WESAD dataset, physiological and motion signals captured by the two sensors are filtered and downsampled to the same frequency, followed by a 10-second non-overlapping segmentation as proposed by \cite{ssl_ecg}. Finally, the SPD matrix series can be calculated from multimodal data segments. For BP4D+ dataset, we first calculated the Euclidean distance between each of the two facial landmarks for each video frame using the provided 2D/3D/thermal facial landmarks, and automatically selected the 10 most discriminative distances by feature selection based on Anova F-value. Then the distance vector and the synchronized physiological signal vector were concatenated together to form a new augmented vector. Subsequently, all the augmented vectors are sliced into 1-second non-overlapping data segments. In the end, the obtained SPD matrix sequences can be extracted from prepared data segments. Fig. \ref{Fig2:spd matrix} shows 
an SPD matrix of a \textit{pain} class sample in BP4D+ dataset that generated from the 10 facial distances between 2D landmarks and 8 physiological signals. During the pain task, the subject was asked to immerse hands in ice water, and her mouth was involuntarily opened, hence the distances automatically selected were all based on the landmarks in the lip region, which is consistent with findings in the literature that lip movements such as oblique lip raising\cite{patrick_obliquelip}, horizontal lip stretching\cite{leresche_lipstretch}, etc. are related to pain. Both intra-modal and inter-modal correlations 
can be observed from the SPD matrix. Among all physiological signals, there was a strong correlation between electrodermal activity (EDA) and respiration signal (respiration rate (RR) and respiration volt (RV)) when the subject was suffering from pain. Among the selected distances, the 2nd, 3rd and 6th distances were more correlated with each other. Furthermore, the association between physiological and facial indicators can be explored, i.e. EDA, RR and RV were also positively correlated with the 9th and 10th distances.
\begin{table*}[h]
\caption{Stress detection performance on WESAD dataset using the sample covariance matrix S, the cross-covariance matrix C and the proposed SPD representation P. }

  \centering
\setlength{\tabcolsep}{0.8mm}{
\begin{tabular}{ccccccccccccc}
\toprule\midrule 
&  \multicolumn{4}{c}{Wrist Sensor} &
\multicolumn{4}{c}{Chest Sensor} & \multicolumn{4}{c}{Wrist+Chest Sensors}\\
\cmidrule(lr){2-5}\cmidrule(lr){6-9}\cmidrule(lr){10-13}

Modalities& \multicolumn{2}{c}{Physio + Motion} & \multicolumn{2}{c}{Only Physio} &
\multicolumn{2}{c}{Physio + Motion} & \multicolumn{2}{c}{Only Physio} &
\multicolumn{2}{c}{Physio + Motion} & \multicolumn{2}{c}{Only Physio}
\\\cmidrule(lr){2-3}\cmidrule(lr){4-5}\cmidrule(lr){6-7}\cmidrule(lr){8-9}\cmidrule(lr){10-11}\cmidrule(lr){12-13}
     Features     & Accuracy  & F1score   & Accuracy  & F1score & Accuracy  & F1score & Accuracy  & F1score & Accuracy  & F1score & Accuracy  & F1score\\\midrule 
     
with S (baseline)    & 89.23 & 88.07 & 90.13 & 89.73 & 90.64 & 90.49 & 90.66 & 89.02 & 92.88 & 92.78 & 93.06 & 93.00 \\
with C (baseline)    & 90.91 & 90.04 & 88.07 & 87.28 & 91.30 & 90.94 & 89.42 & 88.44 & 93.12 & 92.73 & 93.08 & 91.84 \\\midrule 
with P (m = 2)    & 94.08 &  93.81 & 92.15 & 92.20 & 94.55 & 94.53 & 92.76 & 92.61 & 94.19 & 94.23 & 95.29 & 95.33 \\
with P (m = 3)    & 93.00 & 92.38 & 93.10 & 92.78 & \textbf{95.54} & \textbf{94.76} & \textbf{92.96} & \textbf{92.64} & 94.71 & 93.81 & 95.09 & 94.15 \\
with P (m = 4)    & \textbf{94.65} & \textbf{93.99} & \textbf{93.47} & \textbf{93.25} & 94.25 & 93.97 & 92.12 & 91.75 & \textbf{96.35} & \textbf{95.76} & \textbf{96.88} & \textbf{96.44} \\

\midrule 
\bottomrule
\end{tabular}}
\label{table:wesad1} 
\end{table*}

\begin{table}[h]
\caption{Comparison with State-of-the-art Methods on WESAD dataset (\textit{Stress} vs. \textit{Non-stress})} 
  \centering
\setlength{\tabcolsep}{0.8mm}{
\begin{tabular}{|c|c|c|c|}
\hline\hline
Modality & Methods & Accuracy & F1score \\
\hline
\multirow{ 5}{*}{Wrist} & Schmidt et al. \cite{schmidt_introducing_2018} & 87.12 & 84.11 \\
& Samyoun et al.\cite{samyoun_stress_2020} & 92.1 & 89.7  \\
& Gil-Martin et al.\cite{Gil2022} & 92.7 & 92.55  \\
& Huynh et al.\cite{StressNAS} & 93.14 & -  \\
& Lai et al.\cite{Lai_stress} & 94.16 & 93.62 \\
\cline{2-4}
& \textbf{Proposed Method} & \textbf{94.65} & \textbf{93.99} \\
\hline
\multirow{ 4}{*}{Chest} & Samyoun et al.\cite{samyoun_stress_2020} & 91.1 & 90.2  \\
& Schmidt et al. \cite{schmidt_introducing_2018} & 92.83 & 91.07 \\
& Gil-Martin et al.\cite{Gil2022} & 93.10 & 93.01  \\
& \textbf{Lai et al.\cite{Lai_stress}} & \textbf{96.69} & \textbf{96.61} \\
\cline{2-4}
& Proposed Method & 95.54 & 94.76 \\
\hline
\multirow{ 4}{*}{Wrist + Chest} & Schmidt et al. \cite{schmidt_introducing_2018} & 92.28 & 90.74 \\
& Samyoun et al.\cite{samyoun_stress_2020} & 94.7 & 93.4  \\
& Gil-Martin et al.\cite{Gil2022} & 96.62 & 96.63  \\
& \textbf{Lai et al.\cite{Lai_stress}} & \textbf{97.75} & \textbf{97.74} \\
\cline{2-4}
& Proposed Method & 96.88 & 96.44 \\

\hline
\end{tabular}}
\label{table:wesad2} 
\end{table}

\begin{table*}[h]
\caption{Unimodal and Multimodal pain detection performance on BP4D+ dataset using the sample covariance matrix S, the cross-covariance matrix C and the proposed SPD representation P. (2D/3D/Thermal: 2D/3D/Thermal facial landmarks; Physio: all physiological signals).} 
  \centering
\setlength{\tabcolsep}{0.2mm}{
\begin{tabular}{ccccccccccccccc}
\toprule\midrule 
&  \multicolumn{8}{c}{Uni-modality} &
\multicolumn{6}{c}{Multi-modality} \\
\cmidrule(lr){2-9}\cmidrule(lr){10-15}

Modalities& \multicolumn{2}{c}{2D } & \multicolumn{2}{c}{3D } &
\multicolumn{2}{c}{Thermal } & \multicolumn{2}{c}{Physio} &
\multicolumn{2}{c}{2D + Physio} & \multicolumn{2}{c}{3D + Physio} & \multicolumn{2}{c}{Thermal + Physio}
\\\cmidrule(lr){2-3}\cmidrule(lr){4-5}\cmidrule(lr){6-7}\cmidrule(lr){8-9}\cmidrule(lr){10-11}\cmidrule(lr){12-13}\cmidrule(lr){14-15}
     Features     & Accuracy  & F1score   & Accuracy  & F1score & Accuracy  & F1score & Accuracy  & F1score & Accuracy  & F1score   & Accuracy  & F1score & Accuracy  & F1score \\

     \midrule 
     
with S (baseline)  & 88.76 & 85.58 & 86.47 & 85.83 & 81.46 & 80.10 & 77.53 & 72.84 & 91.29 & 90.16 & 87.07 & 86.18 & 82.28 & 80.38  \\
with C (baseline)  & 86.96 & 86.43 & 88.84 & 85.85 & 82.60 & 82.11 & 76.70 & 72.41 & 90.49 & 87.72 & 90.26 & 87.36 & 84.24 & 78.56 \\\midrule 
with P(m = 2)    & 91.44 &  89.24 & 91.01 & 88.84 & 82.82 & 82.70 & 81.82 & 81.72 & 92.86 & 90.03 & 92.53 & 90.20 & 89.04 & 88.73 \\
with P(m = 3)    & \textbf{91.59} & \textbf{89.46} & 91.13 & 89.04 & 83.18 & 82.93 & \textbf{83.24} & \textbf{82.42} & 93.04 & 90.85 & \textbf{92.66} & \textbf{90.47} & 88.82 & 88.69 \\
with P(m = 4)    & 90.94 & 88.56 & \textbf{91.27} & \textbf{89.30} & \textbf{83.53} & \textbf{83.37} & 82.81 & 81.42 & \textbf{93.45} & \textbf{91.37} & 92.54 & 90.25 & \textbf{89.07} & \textbf{88.96}\\

\midrule 
\bottomrule
\end{tabular}}
\label{table:bp4d1} 
\end{table*}
\vspace*{-5pt}
\begin{table}[h]
\caption{Comparison with State-of-the-art Methods on BP4D+ dataset (\textit{pain} vs. \textit{non-pain}).} 
  \centering
\setlength{\tabcolsep}{0.6mm}{
\begin{tabular}{|c|c|c|c|}
\hline\hline
Modality & Methods & Accuracy & F1score \\
\hline
\multirow{ 2}{*}{Physiology}& Hinduja et al.\cite{hinduja_multimodal_nodate} & 77.7 & 30.0\\
\cline{2-4}
& \textbf{Proposed Method} & \textbf{81.82} & \textbf{81.72} \\
\hline
\multirow{4}{*}{Vision} 
& Szczapa et al.\cite{Benjamin2020} & 81.86 & 77.34  \\
& Choo et al. \cite{Choo2021} & 84.03 & 83.98 \\
& Huang et al.\cite{Huang2022} & 86.43 & 85.71 \\
\cline{2-4}
&\textbf{ Proposed Method} & \textbf{91.59} & \textbf{89.46} \\
\hline

\multirow{ 7}{*}{Vision + Physiology}  

&${}^{(D)}$Szczapa et al.\cite{Benjamin2020} & 82.77 & 76.32  \\
& ${}^{(F)}$Szczapa et al.\cite{Benjamin2020} & 84.32 & 78.83  \\
& ${}^{(F)}$Huang et al.\cite{Huang2022} & 87.94 & 87.16 \\
& ${}^{(D)}$Choo et al. \cite{Choo2021} & 89.08 & 88.68 \\
& ${}^{(D)}$Huang et al.\cite{Huang2022} & 89.36 & 89.13 \\
& ${}^{(F)}$Choo et al. \cite{Choo2021} &89.80 & 89.46 \\
\cline{2-4}
& \textbf{Proposed Method} & \textbf{93.45} & \textbf{91.37} \\
\hline
\multicolumn{4}{l}{${}^{(D)}$: Decision level fusion, ${}^{(F)}$: Feature level fusion.}

\end{tabular}}
\label{table:bp4d2} 
\end{table}

\subsection{Implementation and evaluation}
All the classification models were implemented using Pytorch. To avoid overfitting, dropout operation was employed after the LSTM layers with a hidden state dimension of 128. The Adam optimizer with a learning rate lr=0.001 was selected to minimize the binary cross-entropy loss function during model training of 50 epoch. Decay coefficients of the first and second moment estimation $\beta_1$ and $\beta_2$ were set to 0.9 and 0.999, respectively. 
In the end, the proposed framework is evaluated using Leave-One-Subject-Out cross validation (LOSO-CV) on WESAD dataset followed by~\cite{schmidt_introducing_2018,StressNAS,Lai_stress} and Subject independent 10-fold cross validation on BP4D+ dataset followed by \cite{hinduja_multimodal_nodate} with two selected metrics: Accuracy and F1score. Both experimental protocols assess the model's capacity to generalize across previously unseen subject data. 

\subsection{Stress detection results on WESAD}
Binary stress detection experiments were performed using the wrist/chest-based data of all subjects in WESAD dataset. The evaluation results of different modality combinations are reported in Table \ref{table:wesad1}. We first verify the necessity of fusing the sample covariance matrix S and cross-covariance matrix C to form the proposed representation P. From the experimental results, it can be observed that the proposed SPD matrix optimizes the detection performance for all modality combinations, compared to those using only matrix S and matrix C. Secondly, to investigate the impact of increasing the dimensionality of the SPD matrix defined in the equation \eqref{eq:5} on the detection results, we performed experiments using representations with different numbers of blocks of S and C (e.g., m=2 meaning that P consists of 2 blocks of S and 2 blocks of C). We found that using more blocks of S and C to compose the proposed representation P slightly enhanced the classification performance. This can be attributed to the increased proportion of cross-covariance information in the high-dimensional SPD matrix, further demonstrating the benefit of correlations between 
multiple modalities at different instants for the classification task. Besides, we only test up to m=4, since we noticed that sometimes the m=3 case performs best, following the trade-off between classification performance and computational cost. Finally, we also explored whether fusing data from different modalities, i.e., motion signals and physiological signals, could lead to a performance gain.
For experiments based on wrist/chest sensor data, combining these two types of data for detection yielded better performance, compared to results based on physiological signals only. When fusing data from both devices (wrist + chest), results using all modalities were not improved, which can be attributed to the redundant information generated by the same motion signals from both devices. When combining all physiological-based modalities, the highest performance (96.88\% accuracy and 96.44\% F1 score) was obtained 
with $P(m=4)$. 
\subsubsection*{\textbf{Comparison with State-of-the-art}}

To validate the effectiveness of the proposed method for fusing motion and physiological information, Table \ref{table:wesad2} shows the comparison results with 5 state-of-the-art methods using multimodal features. For a fair comparison, only methods that use the same experimental protocol were considered.
In the work of Schmidt et al.\cite{schmidt_introducing_2018}, features from the time and frequency domains are used to train a variety of traditional machine learning classifiers, among which the LDA (Linear discriminant analysis) model achieved the best performance. 
Samyoun et al.\cite{samyoun_stress_2020} presented GAN/RNN/MLP-based deep model to generate gold standard chest sensor features from wrist data, and classified emulated features with various machine learning algorithms (e.g., Extra Trees, Random Forest, etc.).
Gil-Martin et al.\cite{Gil2022} proposed a CNN-MLP architecture to extract meaningful features from the Fourier transform (FFT), cube root (CR) and constant q spectral transform (CQT) of signal sub-window. Huynh et al.\cite{StressNAS} used filter bank as model input and automatically selected the optimal model for each modality from 10,000 deep neural networks for training. Finally, features of all modalities were concatenated for classification. Lai et al.\cite{Lai_stress} employed residual-temporal convolution network to process the filtered multimodal signals and proposed various fusion strategies. The above work for comparison simply spliced features from different modalities for prediction and thus ignored the cross-modality correlations.
Overall, our proposed method using the joint SPD representation achieves the state-of-the-art results on wrist sensor data and competitive results on chest/wrist+chest sensor data, respectively, demonstrating its efficiency for integrating multimodal data.

\subsection{Pain detection results on BP4D+}
To further assess the validity of the proposed method, we conducted unimodal and multimodal pain detection experiments on the BP4D+ dataset. The evaluation results are summarised in Table \ref{table:bp4d1}. Similar to the process performed on WESAD, we first consider the results obtained using S or C alone as baseline to explore the importance of combining them in the proposed representation. Based on the results in Table \ref{table:bp4d1}, we reach the same conclusion that the joint SPD representation can improve classification performance. In addition, we can infer that increasing the dimension of P can further boost performance for all modalities. In the end, we also noticed that all the multimodal settings exhibit performance gains compared to unimodal detection results. Among four unimodality (i.e. physiological signal, 2D/3D/thermal facial landmarks), the trained model has the best performance using 2D facial landmarks where recognition accuracy and F1score achieved 91.59\%, 89.46\% respectively. In the multimodal experiments, the best results with accuracy and F1score of 93.45\% and 91.37\% can be observed with 2D + Physio setting. 
\subsubsection*{\textbf{Comparison with State-of-the-art}}
Table \ref{table:bp4d2} shows the comparison results with 4  state-of-the-art methods using 2D facial landmarks and physiological signals.
Due to the diversity of problem formulations and experimental settings, only a few pain detection efforts can be directly compared to our framework. In the work of \cite{hinduja_multimodal_nodate}, a random forest classifier was trained on features consisting of physiological signals and face action units for pain detection. Here we only presented the comparison results based on physiological signals with them, as we did not use AUs for the detection. Our proposed method improves the accuracy by about 4\%. Moreover, our framework achieved a more balanced pain detection with an F1score of 81.72\%. Since most pain detection datasets contain only vision-related information, very little pain recognition work has been carried out based on data from two different domains, i.e. vision and physiology. Therefore, to validate the effectiveness of our proposed approach on fused multimodal data,  state-of-the-art pain recognition methods that accept only visual data were implemented and combined with our physiological signal-based model for comparison. For a fair comparison, only the
facial landmark-based methods were considered. We used the code provided by the authors, and if the code was not available, we followed the parameters provided in their article. Szczapa et al.\cite{Benjamin2020}
represented the facial landmark sequences as trajectories on the  Riemannian manifold. Each point of the trajectory is a Gram matrix computed from the 2D facial landmarks. Then the Global Alignment Kernel (GAK) was used to calculate the similarity matrix between the trajectories, which was used as feature for SVR-based (Support Vector Regression) pain estimation. To compare the classification performance, we replaced the SVR with an SVM (Support Vector Machine). Huang et al.\cite{Huang2022} used a 1D CNN based architecture to extract discriminate features from the normalized distance between 2D facial landmarks for pain recognition. Choo et al. \cite{Choo2021} employed a dual-layer 3D CNN for capturing the spatial-temporal features of the 2D facial landmark sequences. When comparing the performance of pain recognition based on solely visual information, our model performs better as shown in Table \ref{table:bp4d2}. To compare the performance based on multimodal information, we used two fusion techniques that are commonly used in the literature, feature level fusion and decision level fusion, respectively. We first note that the performance of all the vision-based models used for comparison is improved when combined with our physiology-based model, providing side evidence that our model learns discriminative physiological features. Secondly, our model outperforms
other multimodal approaches, both in terms of feature level fusion and decision level fusion, which confirms that the correlation between two modalities is well captured by the proposed method and that inter-modal communication can further contribute to the classification performance. Overall, our method achieves the state-of-the-art results on both unimodal data and multimodal data, validating again its effectiveness.



\section{Conclusion}
In this work, we explore for the first time the feasibility of SPD matrix-based  representations for efficiently fusing physiological and behavioural signals, which can capture simultaneously  correlation information within and across modalities. Tangent space mapping converts the generated SPD matrix time series into linear vectors for its application to the LSTM-based classification. The effectiveness of the proposed method was evaluated on public stress and pain detection datasets. In the end, the proposed framework shows the state-of-the-art results on both stress and pain detection tasks.

\end{document}